\documentclass[twoside,11pt]{article}

\usepackage{sty}
\usepackage{amsmath,amssymb}

\heading{1}{2023}{1-48}{4/00}{10/00}{Mattéo Papin, Yann Beaujeault-Taudière and Frédéric Magniette}

\ShortHeadings{\texttt{MOSAIC}, a comparison framework for machine learning models}{Papin, Beaujeault-Taudière and Magniette}
\firstpageno{1}

\begin{document}

\title{\texttt{MOSAIC}, a comparison framework for machine learning models}

\author{\name Mattéo Papin  \email matpapin0@gmail.com \\
       \addr \'Ecole 42, 75017 Paris, France
       \AND
       \name Yann Beaujeault-Taudière \email yann.beaujeault-taudiere@ijclab.in2p3.fr \\
       \addr Universit\'e Paris-Saclay, CNRS/IN2P3, IJCLab, 91405 Orsay, France \\
       Laboratoire Leprince-Ringuet (LLR), \'Ecole polytechnique, CNRS/IN2P3, 91120 Palaiseau, France
       \AND
       \name Frédéric Magniette \email frederic.magniette@llr.in2p3.fr \\
       \addr        Laboratoire Leprince-Ringuet (LLR), \'Ecole polytechnique, CNRS/IN2P3, 91120 Palaiseau, France}

\maketitle

\begin{abstract}%   <- trailing '%' for backward compatibility of .sty file
We introduce \texttt{MOSAIC}, a Python program for machine learning models. Our framework is developed with in mind accelerating machine learning studies through making implementing and testing arbitrary network architectures and data sets simpler, faster and less error-prone. \texttt{MOSAIC} features a full execution pipeline, from declaring the models, data and related hyperparameters within a simple configuration file, to the generation of ready-to-interpret figures and performance metrics. It also includes an advanced run management, stores the results within a database, and incorporates several run monitoring options. Through all these functionalities, the framework should provide a useful tool for researchers, engineers, and general practitioners of machine learning.\\
\texttt{MOSAIC} is available from the dedicated \href{https://pypi.org/project/ml-mosaic/}{PyPI} repository.
\end{abstract}

\begin{keywords}
  Open-source software, network optimization, classification, regression, Python, neural networks
\end{keywords}

\section{Introduction}

Conceiving optimal machine learning models, especially with artificial neural networks, is known to be a complex art. Although substantial work has been performed, not only on the mathematical understanding of neural networks' (NNs) approximation power \citep{hornik_multilayer_1989,cybenko_approximation_1989,kurkova_kolmogorovs_1991,kurkova_kolmogorovs_1992,rolnick_power_2018}, but also on the heuristic \citep{lin_why_2017,ojha_metaheuristic_2017} and numerical \citep{sagun_universal_2018,li_visualizing_2018} sides, no practically useful bounds, nor any useful recipe, is available in order to a priori construct NNs that achieve a pre-determined degree of performance for a given class of problems. There is no straightforward method to choose the hyperparameters of a neural network to optimize the number of used parameters. In practice, given a machine learning task, only prior experience give clues on which models should be used, and how the set of hyperparameters must be tuned to reach good (let alone optimal) performances. For a simple model like a multi-layer perceptron (MLP) \citep{mlp}, the shape of the network, the number of hidden layers and the number of neurons inside each of them has a great influence on the result.

For more complicated techniques, such as convolutional neural networks \citep{zhang1988shift,lecun_backpropagation_1989,lecun_cnn_1998}, the number of hyperparameters increases as new quantities come to play, such kernel size, padding, stride, pooling function, etc. Very often, the strategy consists in choosing enormous networks, typically using millions of parameters, to solve the problem. Although such large number of parameters is probably too much in most situations, this gives an insurance of bringing enough expressive power to tackle the complexity of the targeted problematic.

\begin{figure}[ht!]
  \centering
  \includegraphics[width=\textwidth]{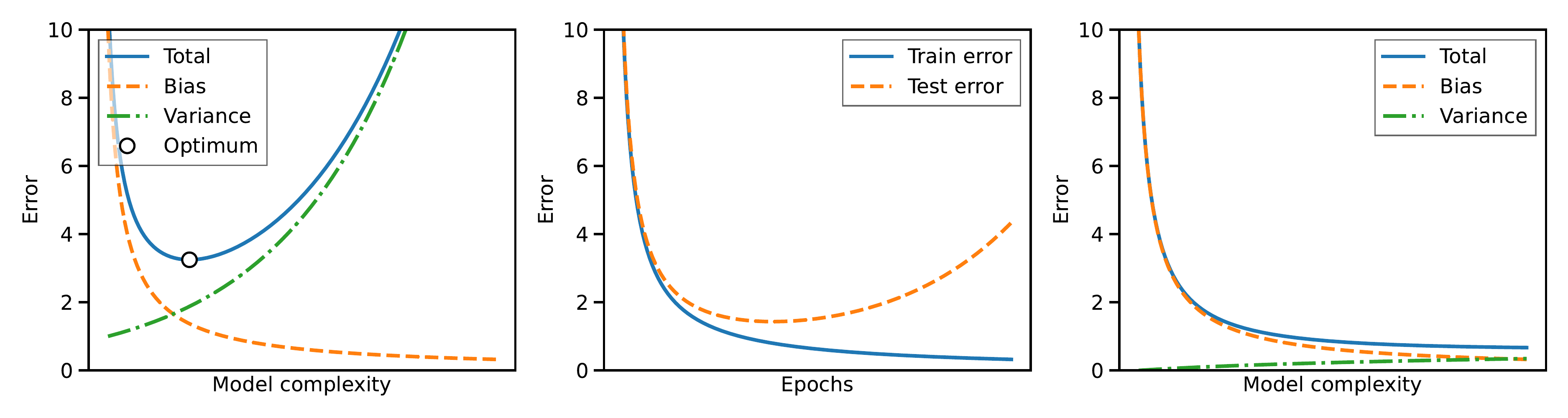}
\caption{Schematic illustration of the performance of NNs' losses according to the relative complexities of the model and data. Left: bias-variance tradeoff, typical of a situation where data complexity $\gg$ model complexity. Middle: overfitting (model complexity $\gg$ data complexity). Right: well-balanced network and data (model complexity $\sim$ data complexity).}
\label{fig:variance}
\end{figure}

The problem of this ``brute-force'' approach is that it requires a lot of labeled input data in order to truly constrain the parameters of the model. Furthermore, this introduces the need for a lot of computing resources along the training process and the use of the model. In addition, the difficulty of optimizing a neural network increases dramatically with its size, further augmenting the computational resources in order to attain satisfactory performances. As shown on the left part of Figure \ref{fig:variance}, when a statistical model is trained on data, two kinds of error occur. These are driven by the respective complexity of the data and the model. If the model is simpler than the data, i.e., the number of parameter is smaller than what is required in order to satisfactorily learn the data, a bias error is pre-eminent, due to the model underfitting the data by learning only its roughest features. On the opposite, in the case of too complex models, a new type of error appears, the so-called variance error. It corresponds to the parameters being too numerous with respect to the data size, so that they end up learning irrelevant features of the training data set, such as noise or unnecessarily moments of the data distribution, which in turn deteriorates the generalization power of the network, causing the test losses to rise. An optimal model is in the middle, where the two kinds of errors balance and result in a minimal total error. This is called the bias-variance tradeoff \citep{bvtd}.

Thus, the problem of using models that are too large with respect to the data is the induction of large variance of the validation loss. The effect of this overfitting can be seen in the middle panel of Figure \ref{fig:variance}. 

Most of the modern implementations fight against overfitting by using a tremendous quantity of data to lower the variance. As seen at the right of Figure \ref{fig:variance}, in that case, the variance becomes negligible and, provided the model is sufficiently expressive, it can effectively handle the complexity of the problem.

In a context where the resources are limited, like when porting a network on reconfigurable electronics (FPGA) or on huge data sets that requires long treatment times, it can be interesting to tune carefully the network in order to reduce the number of parameters without reducing the global performance. 
Furthermore, in the case of more sophisticated techniques, like graph convolutional networks (GCN) \citep{GCN}, another key problem is present. Namely, that there are plenty of different algorithms available \citep{survey_gcn}, all of them having properties that can hardly be predicted accurately. In that case, it is interesting to test extensively the different algorithms to choose the best one for a dedicated problem. 

In both cases, it is highly interesting to test a large collection of models on a data set and to compare their performance. This is a subject that is well-covered by the machine learning community. Plenty of documentation explains how different classes of models compare and their specificities \citep{}. But the problem of most of the approaches consists in their ``one-shot'' aspect. Indeed, a set of concurrent models are tested and compared at the same time, and when a new model or data set is added to the collection, all the corresponding tests have to be run again.

In this work, we try to ease to the maximum the solution of such problems, with in view a long-term approach through the development of useful software. We have developed a framework, called \texttt{MOSAIC}, allowing the sequential test of various architectures on multiple data sets. It provides a database system to store the results and is well-adapted to test hypotheses all along a research project. All the models are evaluated with different sets of hyperparameters, and the run results are uploaded on-the-fly to a database for future comparison. Mainly based on PyTorch \citep{torch}, it is very generic and can be adapted almost effortlessly to any class of model and data. Our framework also provides several tools to compare effectively the models.

\section{The \texttt{MOSAIC} framework}

\begin{figure}[ht!]
  \centering
  \includegraphics[width=.8\textwidth]{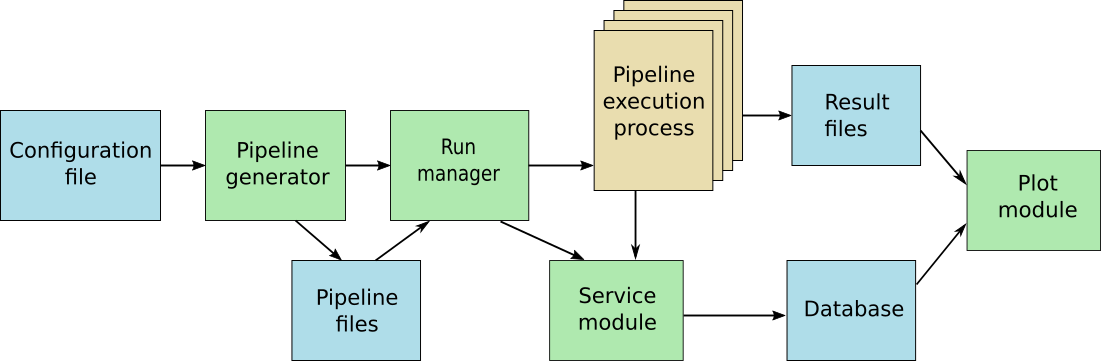}
\caption{Global scheme of the \texttt{MOSAIC} framework. Pipelines are generated from a configuration file. These pipelines are executed in parallel, and the results of the training are stored in files and in a database. In a second step, an analysis module queries the database to produce the comparison plots. }
\label{fig:plan}
\end{figure}

\texttt{MOSAIC} is an open-source Python framework which provides all the facilities to test extensively machine learning models on arbitrary data. It is based on the concept of pipelines, which fully represent a treatment chain including data, data formatting, and model training, all of which possibly including different parts. All the hyperparameters are also included in the pipeline. They are generated by a subpart of the system, via interpreting a configuration file. Once generated, they will be executed, possibly in a parallel way. A standard training procedure is applied using an Adam optimizer \citep{kingma_adam_2017}, and at every step, a training and testing error is computed. At the end of the training, the performance and different indicators are stored in the database. The model parameters and the learning curves are stored in output files. During all the execution process, an advanced error monitoring is performed. Each pipeline can easily be executed multiple times to fill the database with different models and different hyperparameters values. At any time, the analysis module can be used to extract the desired information from the database and produce comparison plots. Figure \ref{fig:plan} shows the entire workflow of the framework.

\section{Pipelines generation}\label{sec:pipelines_generations}

The pipelines are generated from a configuration file (see \ref{app:config_file_pipelines}). We use the \texttt{ini} format, where parameters are written in text within a section indicated by a word between square brackets. Two mandatory sections describe the different parameters of the system itself.

The first one, named \texttt{MONITOR}, includes all the parameters for the run manager, including the requirement for GPUs, the level of parallelism, the cache path and size and the multiplicity. This last parameter allows executing multiple instances of the same pipeline in order to collect statistics. 

The second one, named \texttt{PROCESS}, gives the hyperparameters of the training: initial learning rate, loss function, and so on. It also contains the paths to the classes implementing custom metrics, such as user-defined loss functions. Two crucial elements of this section are the data and model schemes describing the pipeline. They are a sequence of actions that respectively refer to the data set building pipeline and the model. For example, we could have a sequence \texttt{``Data loading | enrichment | normalization''} for the data set. Additionally, one may combine simple models into more complex ones. For example \texttt{``Convolution1 | pooling | Convolution2 | polling | readout | mlp''}. The different parts of the pipeline scheme must be implemented as classes and have a dedicated section.

As an illustration, a section for a funnel-shaped MLP can be represented by
\begin{verbatim}
[mlp_funnel]
type = mlp
class = mlp_funnel
path_to_class = ./mosaic/share/mlp.py
length = 4
width = {2-4},8
\end{verbatim}

The type must be the name present in the scheme. The class and path indicate what class to import. The other parameters are specific to the class. In the simplest configurations, the values are a single number or string. It can also be, as is here the case for the ``width'' parameter, a range of values. In that case, multiples pipelines will be generated, that correspond to all the different values allowed by the range (here, 2, 3, 4, and 8). One must be cautious when giving ranges to multiple parameters, because each cross-possibility among the different class-specific parameters will generate a pipeline, and their total number can grow quite large if several wide ranges are used simultaneously. Thus, for large-scale exploratory studies, it is desirable that the users already have some insight on the relevant ranges of each parameter, or coarse-grain the range of explored values, possibly iteratively refining them after assessing the performances with the help of the plotting tools provided by \texttt{MOSAIC} (see Sec. \ref{sec:results_analysis}). All the generated pipelines are stored in \texttt{JSON} files to be executed directly or later.

A convenient corollary of structuring the runs into pipelines is that it results in minimal length, non-ambiguous characterization of a run. This is particularly useful for generating the plot labels. For instance, the four funnel-shaped MLPs defined in the example \texttt{[mlp\_section]} can be represented as \texttt{mlp\_funnel(4,2)}, \texttt{mlp\_funnel(4,3)}, \texttt{mlp\_funnel(4,4)} and \texttt{mlp\_funnel(4,8)}.

\section{Run and performance indicators}

When the pipelines are generated, they can be run by the core of the framework, the run manager. Every element of the pipeline must be implemented by a Python class providing the required methods. The class API is very similar to the \texttt{Torch.nn.module} class. A forward method provides the evaluation of the data by the model, and a backward method (which can be implicit) is used for computing the partial derivatives. Another method should provide the parameter values of the model. Two methods implement the saving and loading of the model in a file (it can be any format, not only a PyTorch tensor). A last method provides information to the other elements of the pipeline.

To execute a pipeline, every element is called sequentially, and a communication between them is organized following two modalities. First, the result of every element is sent directly to the second. There is no constraint on the format of this return value, except for the last element of the data set model, which must provide a training and testing data loaders (in PyTorch format).

Another way of communication between the elements is based on the \texttt{info} method. Any class can, via this method, output  a dictionary of valued parameters which is made available to every subsequent classes, especially for initialization. For example, the data set class should provide the shape of the inputs for the initialization of the neural network input layer. The names of the class objects must be standardized across the classes entering a pipeline.

During its execution, the model defined by a given pipeline is trained following a traditional training loop, led by an Adam optimizer. Every pipeline is an independent process, communicating with a service module to access to the different services (cache, database, etc.). The run manager follows the progression of the processes and launches new ones as soon as computing resources are available.

At the end of the training loop, various performance indicators, which we detail in Subsection \ref{subsec:perf_indicators}, are calculated and stored in the database for further analysis.

\subsection{Performance indicators}\label{subsec:perf_indicators}

We have implemented six performance indicators in the framework. Each of these provides a valuable, global performance metric, and their collection can be used to efficiently compare the models. In the following, the total number of epochs will be denoted $n$, a given epoch will be indexed by the integer $e=1, \dots, n$, and the loss functions at the epoch $e$ will be written $L_\text{train/test}(e)$.

The first two metrics are the final test and train losses. These are the  traditional, and usually considered most important, indicators to compare the performance of different models. Due to the random nature of the parameters' initialization and the stochastic nature of the Adam optimizer, it is often interesting to try multiple times the same pipeline. This allows users to acquire statistics over the (data, model) couple, as allowed by the \texttt{multiplicity} parameter described earlier. When the multiplicity of a given pipeline is not too large, it is often sufficient to compare only the train and test losses across its different executions in order to identify well-performing runs.

The third indicator we have implemented measures the overfitting of the model. We have chosen the difference of the last train and test error, normalized by the global range of these errors:

\begin{equation}
  \text{overfitting} = \frac{\mid L_\text{test}(n) - L_\text{train}(n) \mid }{\max\left(L_\text{train/test}(n)\right) - \min\left(L_\text{train/test}(n)\right)}.
\end{equation}

The smaller the indicator is, the lower the overfitting.

Fourth, the framework also implements a convergence rate indicator, which gives clues on whether the convergence is reached after the training. For this, we consider the last ten percents of the training loss curve, and carry out two operations. First, the gradients over these points is calculated using a first-order finite difference, and the average gradient is computed as a difference of averages:

\begin{align}\label{eq:slope_def}
    \text{slope}(e) &= L_\text{train}(e)-L_\text{train}(e-1),
    \\
    \overline{\text{slope}} &= \frac{1}{\lceil n/20 \rceil} \left\{\sum_{e=n-\lceil n/20 \rceil}^{n} L_\text{train}(e) - \sum_{e=n-\lceil n/10 \rceil}^{n-\lceil n/20\rceil} L_\text{train}(e) \right\}.
\end{align}

The slopes thus obtained give a rough idea about the convergence. Averaging over the two halves of the last ten percents smoothens sudden jumps of the loss that might occur from one epoch to the following one, which ultimately increases the quality of this metric. In addition, it guarantees that the slope is asymptotically vanishing, which reflects that beyond a certain number of epochs, it is reasonable to expect that the networks' performances will on average stay constant up to negligible variations. Second, since the gradients altogether incorporate information about the possible fluctuations of the loss during the last epochs, we split them into two parts, according to whether they are larger or smaller than the average. The first (resp. second) ones correspond to relative increases (resp. decreases) of the loss, and are thus counted as statistical biases of the final slope towards larger (resp. smaller) values. For each subset, the standard deviation is calculated: 

\begin{align}\label{eq:slope_stds_def}
    \sigma_+ &= \sigma(\{\text{slope}(k) \text{ such that } \text{slope}(k) \geq \overline{\text{slope}} \text{ and } k \geq n - \lceil n/10 \rceil\}), \\
    \sigma_- &= \sigma(\{\text{slope}(k) \text{ such that } \text{slope}(k) \leq \overline{ \text{slope}} \text{ and } k \geq n - \lceil n/10 \rceil\}). 
\end{align}

This indicator is finally presented as $\overline{\text{slope}}^{\sigma_+}_{\sigma_-}$. A negative value indicates that the loss function is likely to decrease if a few epochs are added, whereas a positive or very small slope hints that the model is unlikely to improve and might have reached a local minimum. In the former case, a re-run mechanism allows continuing the training by restarting the corresponding pipeline from the last obtained results.

Fifth, we have implemented a trainability indicator. The idea is to measure how fast a given instance of the model has been trained on the data set. For this purpose, we consider the integral of the train loss curve with respect to the x-axis:

\begin{equation}
  \text{trainability} = \sum_{e=1}^{n} L_{\text{train}}(e).
\end{equation}

If the integral is small, it means that the model has learned fast. Conversely, a high trainability hints that the model is learning the data slowly. This criterion can be used to choose between two models (or classes thereof) otherwise presenting the same performances. If the trainability is lower for one of them, it has been easier to train. 

Following the same idea of selecting equally performing models, a last indicator is simply the training time. This indicator is often of lesser importance since it does not reflect the bare performances of the models; moreover, it depends on the machine running the training. If the test is performed on a single machine or in a homogeneous environment, then this value bears more relevance, and can be used in contexts where the full execution of the pipeline is critical, e.g., when the quantity of data to treat becomes a limiting factor. 

\section{Result analysis}\label{sec:results_analysis}

All the results of the runs are stored in a database. We have chosen the \texttt{SQLite3} file format, in order to benefit from the power of the SQL language and the flexibility of the file storage. SQL allows us to be very accurate on the choice of the models that participate in our studies, up to the point one can easily and quickly query batches of data or cherry-pick specific runs.

\begin{figure}[ht!]
  \centering
  \includegraphics[width=0.8\textwidth]{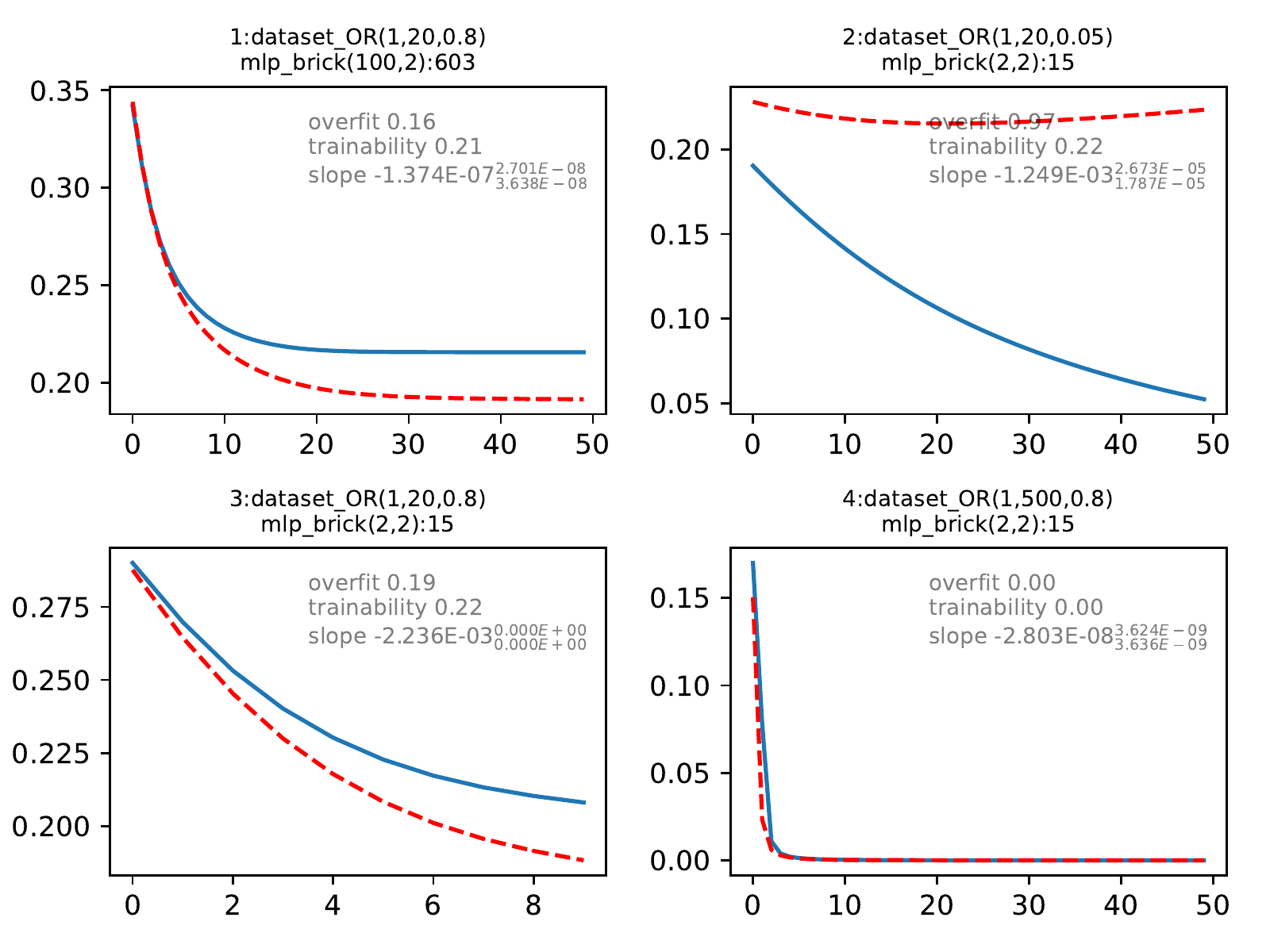}
\caption{To check the convergence of the models, a PDF file can be generated containing all the configuration curves of the different selected models. The different indicators are in the legend (over-fitting, trainability and slope). The train and test loss curves are represented by the solid blue line and dashed red line, respectively. With this format and specific examples, one quickly identifies that run 1 converges to too high values, run 2 overfits, run 3 is likely not converged yet, and run 4 shows excellent convergence.}
\label{fig:lossplots}
\end{figure}

The analysis modules provide two kind of plots. The first one is a collection of train and test losses plots along the epochs, as represented on Figure \ref{fig:lossplots}. Each plot also includes the performance metrics introduced in \ref{subsec:perf_indicators}, except for the runtime. The plots can be selected by a SQL request. By default, all the plots are grouped in a single PDF file, but the number of pages can be very important if the study includes numerous models. In particular, overfitting and convergence indicators can be used to select problematic trainings.

\begin{figure}[ht!]
    \centering
    \includegraphics[width=.49\textwidth]{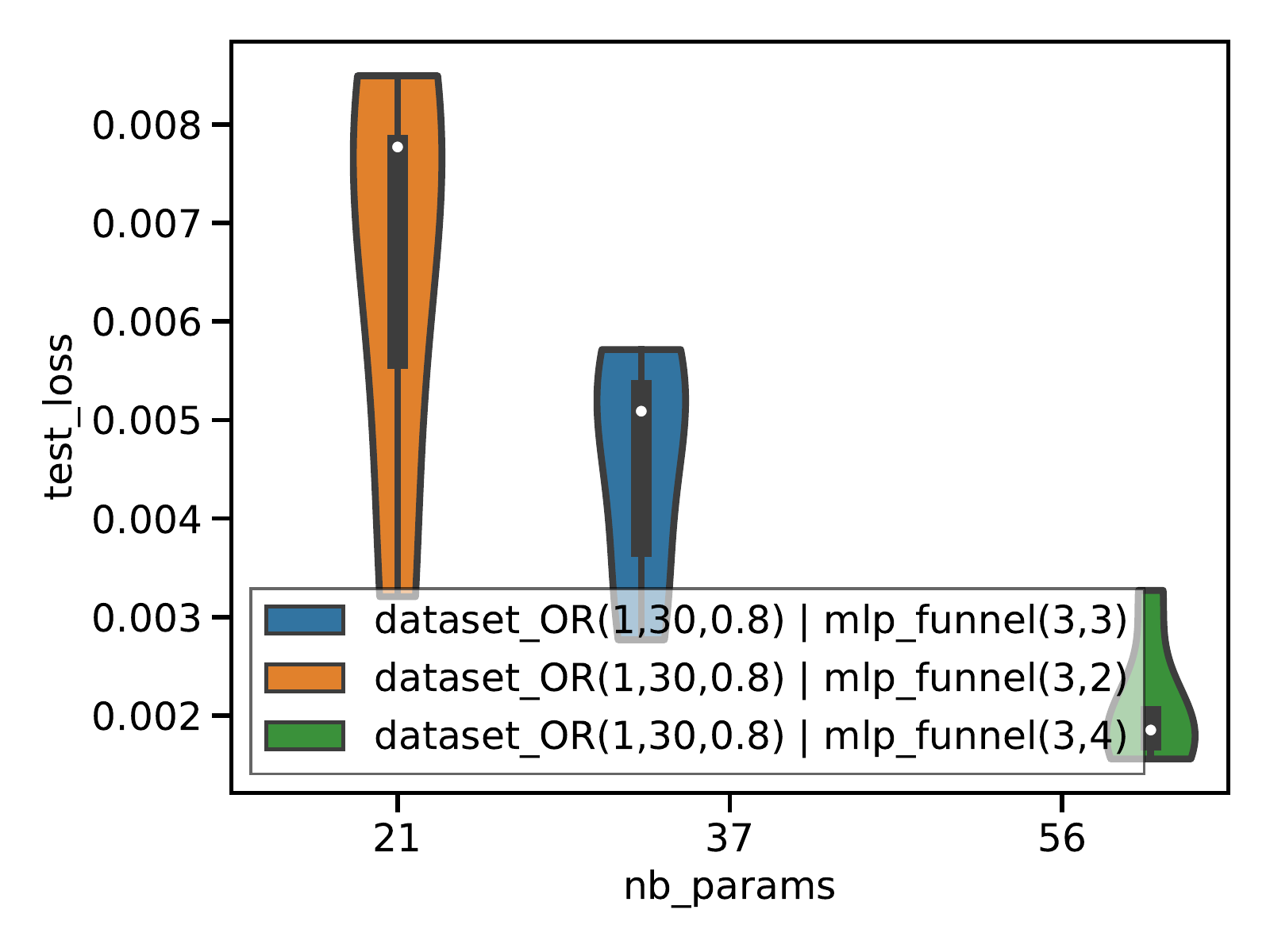}
    \includegraphics[width=.49\textwidth]{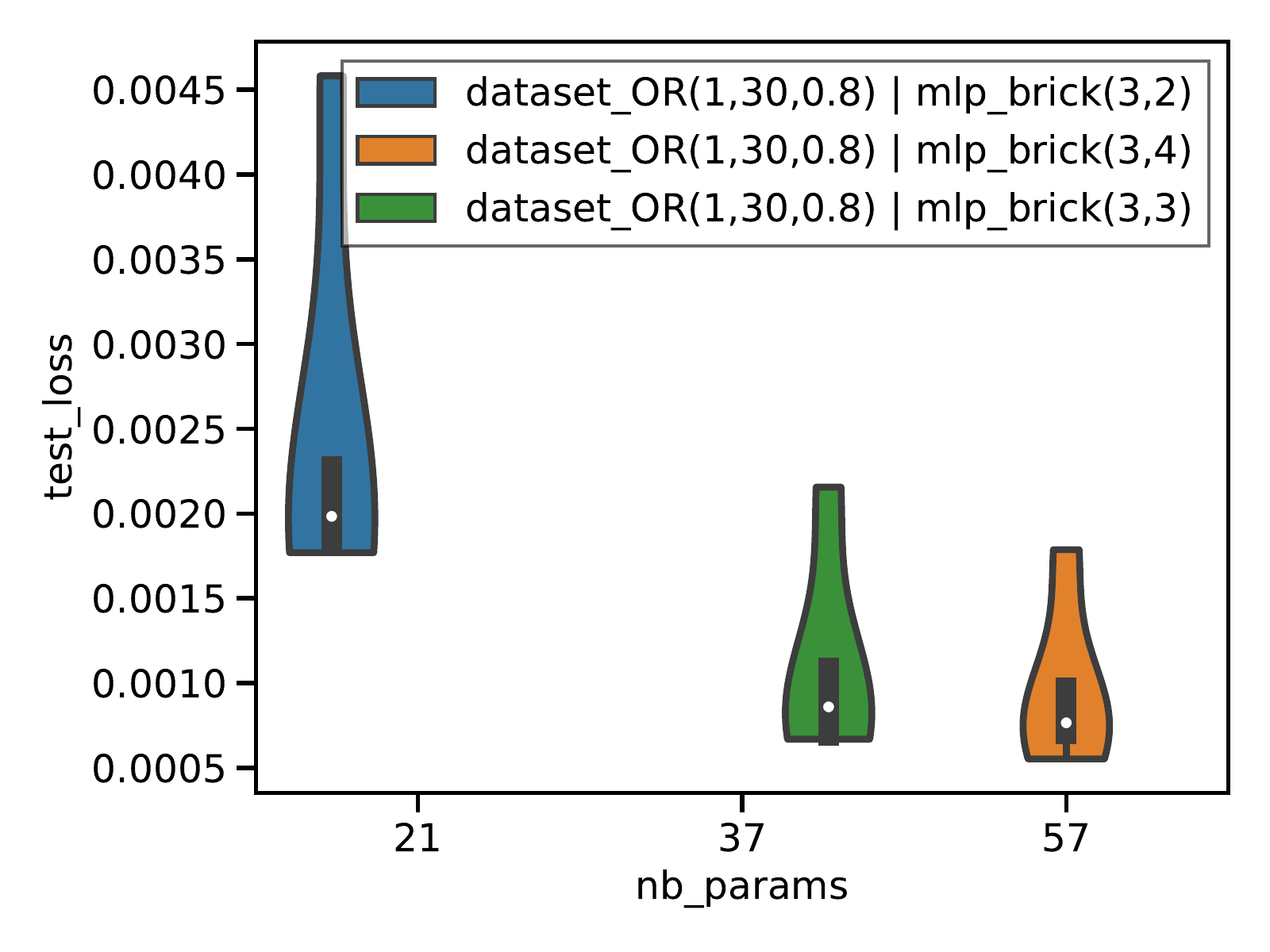}
    \caption{Example of comparison plots (``meta-plot''). The abscissa is the number of parameters of the model, and the ordinate is the error made on the test data set. Such plots allow quickly comparing in detail how the performances of a given class of models, here funnel-shaped (left) and brick-shaped (right) MLPs, evolve with the number of parameters.}
    \label{fig:metaplots_violins}
\end{figure}

The other kind of plot produced by \texttt{MOSAIC} is called meta-plots. They are composed of different curves representing any indicator in ordinate and any parameter in abscissa. Figure \ref{fig:metaplots_violins} shows an example of such format, in the form of violin plots. In such plots, the discrete results condition a Gaussian kernel density distribution \citep{hastie_generalized_1990}, which allows estimating a continuous probability density function from a finite number of runs. This presentation allows displaying all the available information of a pipeline's results, making it extremely convenient to visually and finely compare different hyper-parametrizations. However, probability distribution are poorly suited for comparison of a large amount of models at the same time. 

\begin{figure}[ht!]
    \centering
    \includegraphics[width=.49\textwidth]{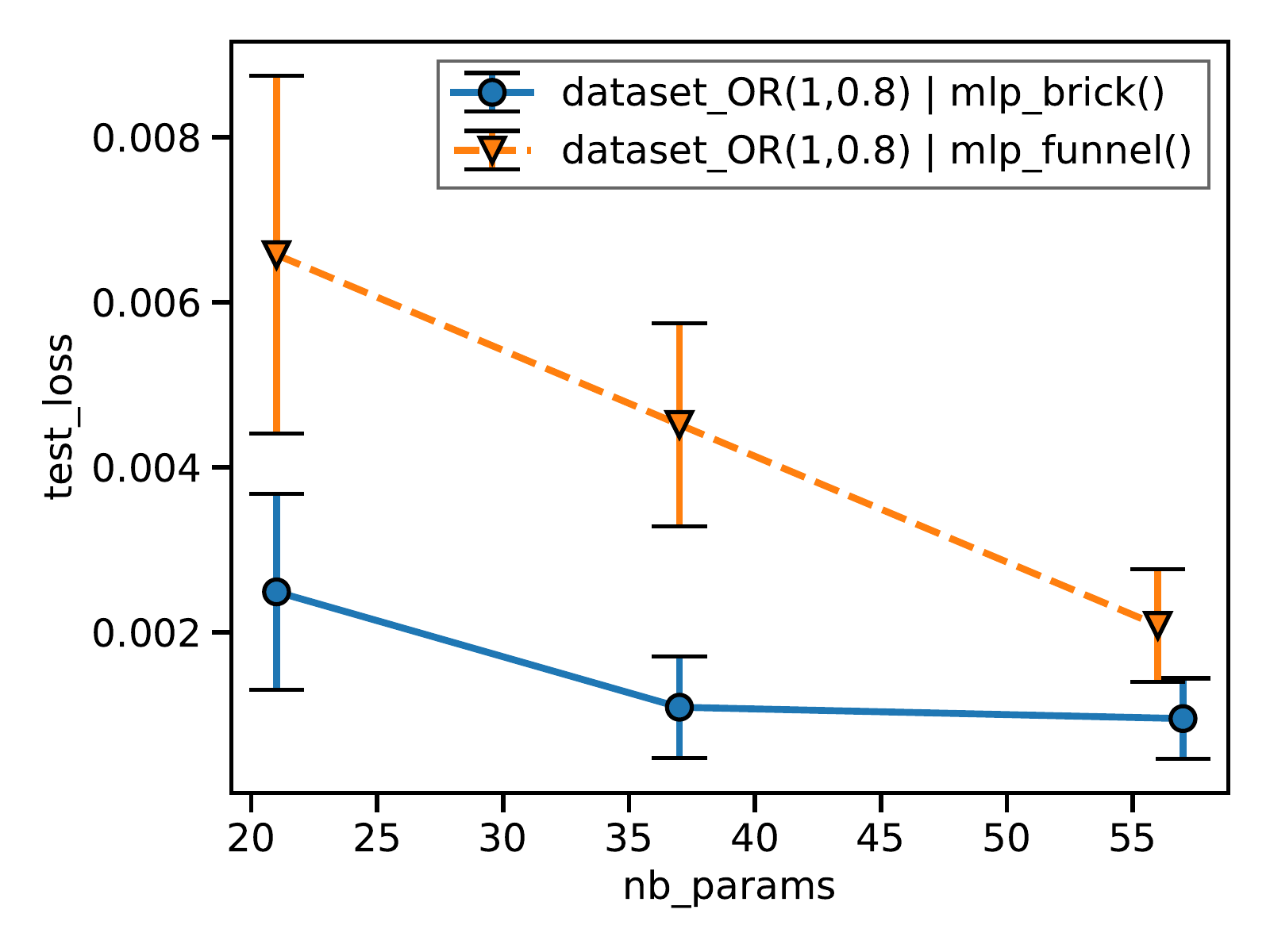}
    \includegraphics[width=.49\textwidth]{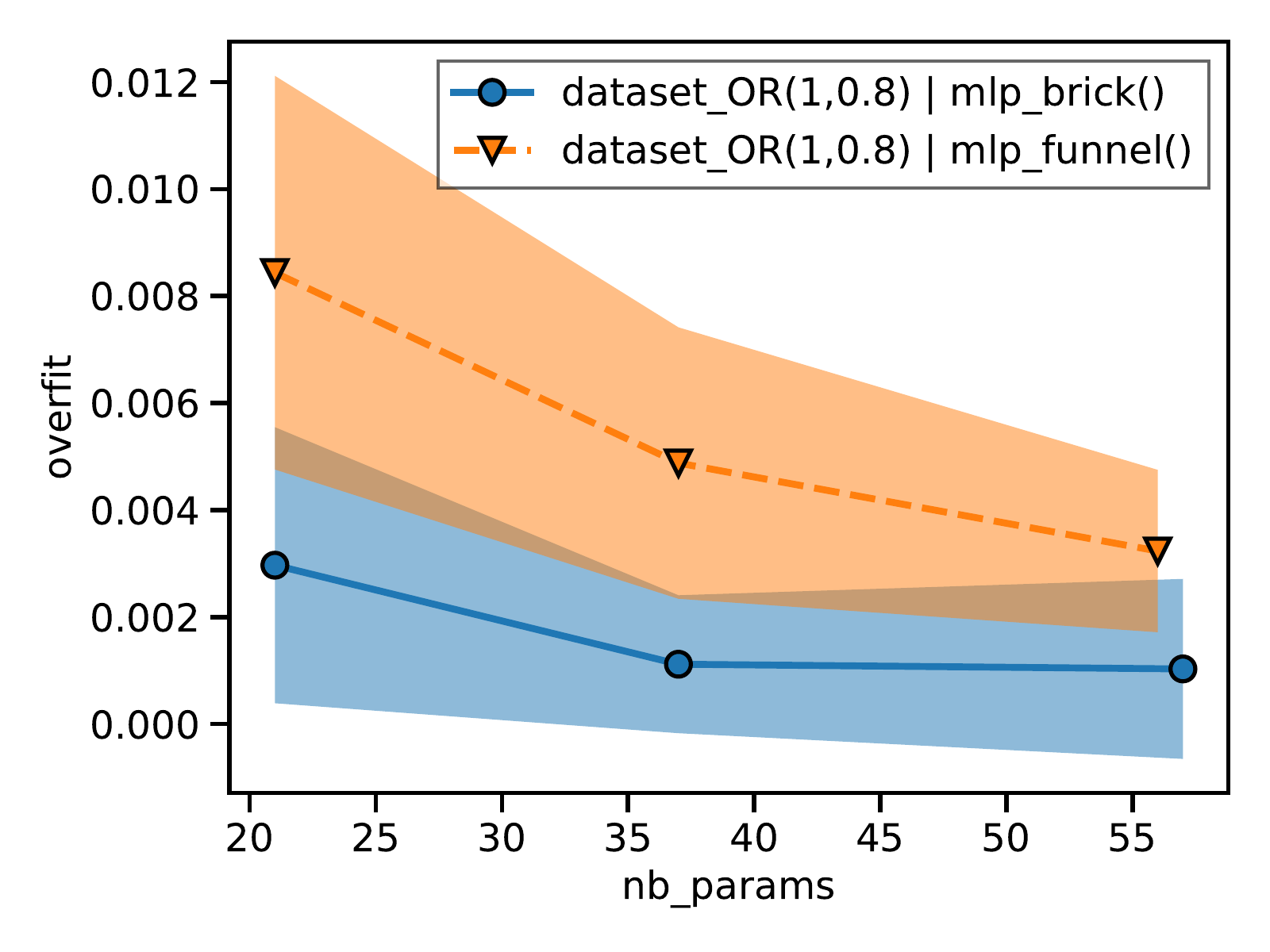}
\caption{Example of comparison plots. The abscissa is the number of parameters of the model, and the ordinate is the error made on the test data set (left) and the overfit (right). The style of the error (left: bars, right: filled) is controlled by a parameter in the \texttt{.ini} configuration file. Note that in the captions, the section of the pipeline describing the model's length and width is automatically adjusted to the plotted data.}
\label{fig:metaplots_lines}
\end{figure}

When one wants to compare several kinds of models, line plots are particularly useful, as each curve condenses the most information into single data points. Figure \ref{fig:metaplots_lines} depicts two examples of such meta-plots, comparing here two otherwise identical MLPs that differ only by their shape (funnel versus brick). Here, each point is an average over all strictly identical pipelines, whereas the error bars represent one standard deviation.  As illustrated on this figure, it is often interesting to choose the number of parameters as abscissa, in order to compare the quality of the models depending on their inherent complexity. Indeed, the number of parameters is a good indication of the needed time for training and of the needed resources to implement the network on a resource-constrained environment (FPGA or neuromorphic chips, for instance). As an illustration, one quickly reads from Fig. \ref{fig:metaplots_lines} that for this data set, brick-shaped MLPs clearly outperform the funnel-shaped ones.

To get meaningful results from these plots, it is often useful to group different models, or different values of hyperparameters, when studying their performance. This allows a finer selection of the hyperparameters and their values. To do this, a special parameter, named \texttt{key}, can be added in the meta-plots configuration file (see \ref{app:config_file_metaplots}). The keys are regular expressions that are stored in the database. This structure eases the grouping by simplifying the SQL request. In Figure \ref{fig:metaplots_lines}, for example, the two plots are obtained by displaying on a single figure pipelines corresponding to different depth and shapes.

The implemented indicators can be inoperant in very specific conditions. In order to allow any specific case, it is possible to choose any field from the database that has been filled by any class during the call of the info method. This way, any specific analysis can be performed without restriction of the predefined indicators. 

\section{Installation and use of the \texttt{MOSAIC} program}
MOSAIC is freely available as a Python package from the PyPI package
manager \citep{pypi}, which eases its installation. The PyPI page of the project is \href{https://pypi.org/project/ml-mosaic/}{https://pypi.org/project/ml-mosaic/}. The framework provides a complete text interface for each functionality,
including pipelines generation, run start and monitoring, re-running of
incomplete trainings and graphics generation (basic and advanced). A
full documentation of the interface and its multiple parameters, and a tutorial explaining step by step how to set up a comparison experiment,
are available on the project's web page, \href{https://llrogcid.in2p3.fr/the-mosaic-framework/}{https://llrogcid.in2p3.fr/the-mosaic-framework/}. Both are also present into the downloaded package, as doc.md and
tutorial.md.

\section{Conclusion and perspectives}

\texttt{MOSAIC} is an open-source framework designed to carry out compared performance studies on machine learning models. It provides all the necessary tools to integrate the testing logic in the long term of a feasibility study. It also includes advanced tools to compare the models and select the optimal ones. Such a key-in-hand framework should be useful to a wide range of users, ranging from novices to experimented practitioners, by drastically reducing the amount of time spent in developing and benchmarking their machine learning programs.

In future releases, an important improvement will be to support the use of clusters through calls to batch systems. We are presently working on the integration of the condor \citep{condor} system to distribute the process over a GPU-equiped cluster. It should speed up considerably the global process.

% Acknowledgements should go at the end, before appendices and references

\acks{The authors acknowledge the support of the French Agence Nationale de la Recherche (ANR), under grant ANR-21-CE31-0030, project OGCID. The authors acknowledge financial support from the P2IO LabEx (ANR-10-LABX-0038).}

% Manual newpage inserted to improve layout of sample file - not
% needed in general before appendices/bibliography.

\newpage

\vskip 0.2in
\bibliography{biblio}

\newpage

\appendix
\section{Example of configuration files}\label{app:config_files}

The following two sections give exhaustive examples of configuration files.

\subsection{Pipeline configuration file}\label{app:config_file_pipelines}
\begin{verbatim}
[PROCESS]
lr = 1e-2
epochs = 200
loss_function = MSELoss
data_scheme = dataset_generator, data_augmentation, data_transformation
pipeline_scheme = convolution, readout, mlp
run_files_path = .runs

[MONITOR]
need_gpu = True
gpu_available = cuda:1, cuda:2
nb_processus = 8
multiplicity = 4
cache_database_path = ./cache.db
cache_size = 1G

[dataset_gen]
type = dataset_generator
class = dataset_generator
path_to_class = ./path/to/dataset/dataset_generator.py
batch_size = 64
data_size = 10000
train_prop = 0.9
key = data_{class}

[data_augmentation]
type = data_augmentation
class = data_augmentation
path_to_class = ./path/to/data_augmentation/data_augmentation.py
data_size = 1000
key = data_{class}

[data_transformation]
type = data_transformation
class = transform
path_to_class = ./path/to/data_transformation/data_transformation.py
style = normalisation, standardisation
key = data_{class}

[convolution]
type = convolution
class = convolution
path_to_class = ./path/to/convolution/convolution.py
pooling = True, False
nb_convolution = {3, 5}, 8
key = conv_{class}

[readout_1]
type = readout
class = simple_readout
path_to_class = ./path/to/readout/readout.py
key = readout_{class}

[readout_2]
type = readout
class = complex_readout
path_to_class = ./path/to/readout/readout.py
key = readout_{class}

[mlp_funnel]
type = mlp
class = mlp_funnel
path_to_class = ./path/to/mlp/mlp_funnel.py
length = 5, {6-8}
width = {2-4}
key = mlp_{class}

[mlp_brick]
type = mlp
class = mlp_brick
path_to_class = ./path/to/mlp/mlp_brick.py
length = 5, 6, 7, 8
width = {2-4}
key = mlp_{class}

\end{verbatim}

\subsection{Meta-plots configuration file}\label{app:config_file_metaplots}
\begin{verbatim}
[global]
file_title = output.plot

[plot_1]
abscissae = nb_params
ordinates = test_loss
include_keys = class, data_size, length
include_values = [dataset_OR], [30, 60], 3
excludes = width, length, data_size
plot_type = line
errorbars_style = bars

\end{verbatim}

\end{document}